\documentclass{article}

\usepackage[utf8]{inputenc} 
\usepackage[T1]{fontenc}    
\usepackage{hyperref}       
\usepackage{url}            
\usepackage{booktabs}       
\usepackage{amsfonts}       
\usepackage{nicefrac}       
\usepackage{microtype}      

\title{The discriminative Kalman filter for nonlinear and non-Gaussian sequential Bayesian filtering}

\author{
Michael C. Burkhart* 
\texttt{michael\_burkhart@brown.edu}
\\
David M. Brandman* 
\texttt{david\_brandman@brown.edu}
\\
Carlos E. Vargas-Irwin 
\texttt{carlos\_vargas\_irwin@brown.edu}
\\
Matthew T. Harrison 
\texttt{matthew\_harrison@brown.edu}
}

\usepackage[boxed]{algorithm2e}
\usepackage{amsmath, amssymb, tikz-cd}
\newcommand{\N}{\ensuremath{\mathcal{N}}}
\newcommand{\RR}{\ensuremath{\mathbb{R}}}
\newcommand{\EE}{\ensuremath{\mathbb{E}}}
\newcommand{\VV}{\ensuremath{\mathbb{V}}}
\newcommand{\SSS}{\ensuremath{\mathbb{S}}}
\newcommand{\tr}{\ensuremath{\text{\tiny T}}}

\begin{document}

\maketitle

\begin{abstract}

The Kalman filter (KF) is used in a variety of applications for computing the posterior distribution of latent states in a state space model.  The model requires a linear relationship between states and observations. Extensions to the Kalman filter have been proposed that incorporate linear approximations to nonlinear models, such as the extended Kalman filter (EKF) and the unscented Kalman filter (UKF). However, we argue that in cases where the dimensionality of observed variables greatly exceeds the dimensionality of state variables, a model for $p(\text{state}|\text{observation})$ proves both easier to learn and more accurate for latent space estimation. We derive and validate what we call the discriminative Kalman filter (DKF): a closed-form discriminative version of Bayesian filtering that readily incorporates off-the-shelf discriminative learning techniques. Further, we demonstrate that given mild assumptions, highly non-linear models for $p(\text{state}|\text{observation})$ can be specified. We motivate and validate on synthetic datasets and in neural decoding from non-human primates, showing substantial increases in decoding performance versus the standard Kalman filter.

\end{abstract}

\section{Introduction}
Consider a state space model, or a hidden Markov model (HMM), for states $Z_t\in \RR^d$ and observations $X_t\in \RR^m$.
The conditional density of $Z_t$ given $X_{1:t}=(X_1,\dotsc,X_t)$ can be computed recursively using the well-known sequential Bayesian filtering equation \cite{chen2003bayesian}:
\begin{equation} 
p(z_t|x_{1:t}) \propto p(x_t|z_t) \int p(z_t|z_{t-1}) p(z_{t-1}|x_{1:t-1}) \; dz_{t-1} \label{e:BF} 
\end{equation}
where the conditional densities $p(z_t|z_{t-1})$ and $p(x_t|z_t)$ are either specified \emph{a priori}, or are learned from training data. The expression is proportional as a function of $z_t$, and the constant of proportionality involves an integral over $z_t$.   

The Kalman filter is based on a special case where, among other things, a linear relationship between states and observations is specified, yielding an analytic solution that can be efficiently computed. However, for most non-linear specifications of $p(x_t|z_t)$, the integral is intractable and requires numerical approximation, such as particle filtering \cite{Gordon:1993, Kanazawa:1995,Doucet:2000}. Moreover, for nonlinear models and high-dimensional observations, filtering often suffers because $p(x_t|z_t)$ is too difficult to learn, regardless of the sophistication of the filtering algorithm.

In this paper, we rewrite equation (\ref{e:BF}) as follows:
\begin{equation} 
p(z_t|x_{1:t}) \propto \frac{p(z_t|x_t)}{p(z_t)} \int p(z_t|z_{t-1}) p(z_{t-1}|x_{1:t-1}) \; dz_{t-1}  \label{e:BF2} 
\end{equation}
To use this filtering equation, we need to learn $p(z_t|x_t)$ instead of $p(x_t|z_t)$, which can lead to dramatic improvements in applications where the dimensionality of the latent states is much less than that of the observed variables. This is especially true in the case of neural decoding, where very high-dimensional neural signals are used to model the evolution of low-dimensional latent-state variables. Besides the striking improvements in learning, this new perspective permits an explicit solution to equation \eqref{e:BF2}, analogous to the Kalman filter, but for models that allow $p(x_t|z_t)$ to be {\em nonlinear} and allow non-Gaussian observations. Moreover, it allows for the inclusion of off-the-shelf discriminative learning techniques (specifically, nonlinear and/or nonparametric regression methods) into fast Bayesian filtering algorithms. We note that our approach is distinct from using discriminative methods for training the standard filters \cite{Abb:2005,hess2009discriminatively}. We call this explicit solution the discriminative Kalman filter (DKF), since it results from partially switching to a discriminative modeling perspective.  

\section{Model specifications}

\subsection{Kalman filter and variants}

The classic, stationary, Kalman filtering model is specified as:
\begin{align} 
\label{e:Z} p(z_t) & = \eta_d(z_t;0,S) \\ 
\label{e:ZZ-1} p(z_t|z_{t-1}) & = \eta_d(z_t;Az_{t-1},\Gamma)  \\
\label{e:KF-X} p(x_t|z_t) & = \eta_m(x_t;Hz_t,\Lambda)
\end{align}
where $A\in\RR^{d\times d}$, $H\in\RR^{m\times d}$, $\eta_d(\cdot;\mu,\Sigma)$ is the $d$-dimensional multivariate Gaussian distribution with mean vector $\mu$ and covariance matrix $\Sigma$, and the covariance $S$ satisfies $S=ASA^\tr+\Gamma$, so that the process is stationary.  Both the linear and Gaussian specifications prove to be poor assumptions in practice, but it is difficult to relax them and retain a computationally efficient, closed-form solution to equation \eqref{e:BF}, which is one of the strong advantages of using a Kalman filter.  

We can relax the linear assumption of \eqref{e:KF-X} in the Kalman, and specify models of the form
\begin{equation} 
\label{e:EKF-X} p(x_t|z_t) = \eta_m(x_t;h(z_t),\Lambda) 
\end{equation}
where $h:\RR^d\to\RR^m$ is nonlinear. The extended Kalman filter (EKF) and the unscented Kalman filter (UKF) \cite{Julier:1997} provide fast approximations to equation \eqref{e:BF} by linearizing $h$ around an appropriate point. The UKF improves the variance approximation in the EKF by evolving a collection of points through the function h. Relaxing the Gaussian assumption, however, usually requires much more computationally demanding algorithms, such particle filtering, which uses Monte Carlo methods to approximate the integrals in equation \eqref{e:BF}. Furthermore, and perhaps more importantly, if the observation model must be learned from data and if the dimensionality $m$ of the observations is large, then most methods perform poorly, including the EKF, UKF, and particle filter, presumably because it is very difficult to learn the function $h(\cdot)$.

\subsection{Discriminative Kalman filter}

Motivated by equation \eqref{e:BF2}, the DKF maintains the modeling assumptions of \eqref{e:Z} and \eqref{e:ZZ-1}, but models $p(z_t|x_t)$ instead of $p(x_t|z_t)$. The model is Gaussian, but in a fundamentally different way, and it is allowed to be nonlinear, i.e.,
\begin{equation} 
\label{e:DKF-X} p(z_t|x_t) = \eta_d(z_t;f(x_t),Q(x_t)) 
\end{equation}
where $f:\RR^m\to\RR^d$ and $Q:\RR^m\to\SSS_d$, using $\SSS_d$ to denote the set of $d\!\times\!d$ covariance matrices. There are several advantages to this approach:
\begin{itemize}
\item There is an exact, closed form solution for $p(z_t|x_{1:t})$ using \eqref{e:BF2} and the component densities specified in \eqref{e:Z}, \eqref{e:ZZ-1} and \eqref{e:DKF-X}. This is true regardless of the functional form of the nonlinearities in $f(\cdot)$ and $Q(\cdot)$. See section \ref{s:exact}.
\item When the dimensionality $m$ of the observations is larger than the dimensionality $d$ of the state space, it can be much easier to learn $f(\cdot)$ in equation \eqref{e:DKF-X} than it is to learn $h(\cdot)$ in equation \eqref{e:EKF-X}. We illustrate this using a real-world neural decoding example in section \ref{s:monkey}.
\item The Gaussian assumption in \eqref{e:DKF-X}, which relates to the states, is often much more natural than the ones in \eqref{e:KF-X} and \eqref{e:EKF-X}, which relate to the observations, particularly when $m\gg d$.  Under mild regularity assumptions, the Bernstein-von Mises Theorem states that $p(z_t|x_t)$ in equation \eqref{e:DKF-X} is asymptotically normal (in total variation distance) as the dimensionality of $x_t$ increases. The observations themselves are not required to be conditionally Gaussian or even continuously-valued. For instance, in neural decoding, the observations are often counts of neural spiking events (action potentials), which might be restricted to small integers, or even binary valued.
\item The DKF subsumes the Kalman filter as a special case by restricting $f$ to be linear and $Q$ to be constant. 
\end{itemize}

\subsection{Exact solution to the discriminative Kalman filter} \label{s:exact}

The modeling assumptions in \eqref{e:Z}, \eqref{e:ZZ-1} and \eqref{e:DKF-X} imply that $p(z_t|x_{1:t})$ is Gaussian, namely,
\begin{equation}\label{e:DKp} p(z_t|x_{1:t}) = \eta_d(z_t;\mu_t,\Sigma_t) , \end{equation}
where $\mu_t$ and $\Sigma_t$ depend on $x_{1:t}$. To see this, fix a sequence $x_1,x_2,\dotsc\in\mathbb{R}^m$, and recursively define $\mu_t=\mu_t(x_{1:t})\in\mathbb{R}^d$ and $\Sigma_t=\Sigma_t(x_{1:t})\in \mathbb{S}_d$ via 
\begin{equation} 
\label{e:DKF} \begin{aligned}  
M_{t-1} &= A\Sigma_{t-1}A^\tr+\Gamma , \\
\Sigma_t & = (Q(x_t)^{-1}+M_{t-1}^{-1}-S^{-1})^{-1} ,  \\
\mu_t & = \Sigma_t(Q(x_t)^{-1}f(x_t) + M_{t-1}^{-1}A\mu_{t-1}) ,
\end{aligned} 
\end{equation}
where $\mu_0=0$ and $\Sigma_0=S$.  Beginning with $p(z_{t-1}|x_{1:t-1})=\eta_d(z_{t-1};\mu_{t-1},\Sigma_{t-1})$, which is true for $t=1$, equations \eqref{e:BF2}, \eqref{e:Z}, \eqref{e:ZZ-1}, and \eqref{e:DKF-X} imply
\[  \begin{aligned}
p(z_t|x_{1:t}) & \propto \frac{\eta_d(z_t;f(x_t),Q(x_t))}{\eta_d(z_t;0,S)} \int \eta_d(z_t;Az_{t-1},\Gamma) \eta_d(z_{t-1};\mu_{t-1},\Sigma_{t-1}) \; dz_{t-1} \\
& = \frac{\eta_d(z_t;f(y_t),Q(y_t))}{\eta_d(z_t;0,S)} \eta_d(z_t;A\mu_{t-1},M_{t-1}) \propto \eta_d(z_t;\mu_t,\Sigma_t) .
\end{aligned} \]
The final step involves a straightforward manipulation of matrices to complete the square. The function $Q(\cdot)$ needs to be defined so that $\Sigma_t$ exists and is a proper covariance matrix.  A sufficient condition that is easy to ensure in practice is $S-Q(\cdot)\in\mathbb{S}_d$.  This proves equation \eqref{e:DKp} and provides an exact, recursive solution for the posterior mean and covariance. 

If $f(\cdot)$ and $Q(\cdot)$ are easy to evaluate and the dimension $d$ of $Z_t$ is not too large, then the DKF can be used in time-sensitive applications.  Additional assumptions can further simplify calculations: for instance, if the conditional variance of $Z_t$ given $X_t$ does not depend on $X_t$, so that $Q(\cdot)$ is constant, then $\Sigma_t$ does not depend on $x_{1:t}$ and can be precomputed in applications. In this case $\Sigma_t$ will quickly converge to the steady-state covariance $\Sigma$ satisfying $\Sigma = (Q^{-1}+(A\Sigma A^\tr+\Gamma)^{-1}-S^{-1})^{-1}$.

\subsection{Modeling philosophy}

Because the DKF separately specifies $p(z_t)$ and $p(z_t|x_t)$ without enforcing agreement between them, there may be no joint distribution of $(Z_t,X_t)$ that has both densities. Nevertheless, equation \eqref{e:DKp} gives a proper distribution for $p(z_t|x_{1:t})$ as long as $\Sigma_t$ is a proper covariance matrix.  This requirement is much weaker than requiring the component densities to agree. Our perspective is that the DKF is directly modeling the posterior $p(z_t|x_{1:t})$ and that this model is defined indirectly via the recursive formulas in equations \eqref{e:DKp}--\eqref{e:DKF}. The component densities provide intuition for how this implicit model works and they also provide guidance for learning the model from training data; see section \ref{s:learning}. The situation is loosely analogous to using an improper prior in Bayesian statistics: as long as Bayes' rule can be formally applied to give a proper posterior, then useful and coherent inferences can still be made.

\section{Learning} \label{s:learning}

The parameters in the DKF are $A$, $\Gamma$, $Q(\cdot)$, and $f(\cdot)$. When learned from data, we use $\hat A$, $\hat \Gamma$, $\hat Q$, and $\hat f$ to denote the learned parameters. 
Perhaps the simplest method for learning the parameters is to learn $A$ and $\Gamma$ from $(Z_{t-1},Z_t)$ pairs using the conditional model for $p(z_t|z_{t-1})$ in equation \eqref{e:ZZ-1} and to separately learn $f$ and $Q$ from $(Z_t,X_t)$ pairs using the conditional model for $p(z_t|x_t)$ in equation \eqref{e:DKF}, in both cases ignoring the overall temporal structure of the data. Since
\[ f(x)=\EE(Z_t|X_t=x) , \]
where $\EE$ denotes expected value, $f$ can be learned using any number of off-the-shelf regression tools, and then, for instance, $Q$ can be learned from the residuals on a held-out portion of the training data. In fact, we think that the ability to easily incorporate off-the-shelf discriminative learning tools into a closed-form filtering equation is one of the most exciting and useful aspects of this approach.

In the experiments below we compare two different nonlinear methods for learning $f$, Gaussian process (GP) regression and neural network (NN) regression. For the NN-regression we learn $Q$ from the residuals after learning $f$. For the GP-regression, we learn $Q$ in two different ways: (i) from the residuals, as usual, and (ii) using an estimate of the posterior variance of $Z_t$ given by the GP-regression framework.  $A$ and $\Gamma$ are learned using linear regression, as is customary when learning the parameters of a Kalman filter from fully observed training data. In all cases, the model parameters are learned from training data and then fixed and evaluated on testing data.  

\subsection{Gaussian process regression} \label{s:GP}

Gaussian process (GP) regression is a popular method for nonlinear regression \cite{Ras:2006}. The idea is to put a prior distribution on the function $f$ and approximate $f$ with its posterior mean given training data. We will first briefly describe the case $d=1$. We assume that we have training data $(Z',X')=(Z_1',X_1'),\dotsc,(Z_n',X_n')$, where $p(z_i'|x_i',f)=\eta_1(z_i'|f(x_i'),\sigma^2)$ and where $f$ is sampled from a mean-zero GP with covariance kernel $K(\cdot,\cdot)$. Under this model
\begin{align} \label{e:GPf}  \EE(f(x)|Z',X') & = K(x,X')(K(X',X')+\sigma^2 I_n)^{-1} Z' \\
\VV(f(x)|Z',X') & = K(x,x)-K(x,X')(K(X',X')+\sigma^2 I_n)^{-1}K(x,X')^\tr \label{e:GPv}
\end{align}
where $K(x,X')$ denotes the $1\!\times\! n$ vector with $i$th entry $K(x,X_i')$, where $K(X',X')$ denotes the $n\!\times\! n$ matrix with $ij$th entry $K(X'_i,X'_j)$, where $Z'$ is a column vector, and where $I_n$ is the $n\!\times\! n$ identity matrix. The noise variance $\sigma^2$ and any parameters controlling the kernel shape are hyperparameters. For our examples, we used the radial basis function kernel (with two parameters: length scale and maximum covariance) and selected hyperparameters via marginal likelihood using the publicly available \texttt{gpml} (www.gaussianprocess.org) package \cite{Ras:2006}.  For our estimate of $f$ we use
\[ \hat f(x) =  \EE(f(x)|Z',X') \]
from equation \eqref{e:GPf}. For later reference, we also define
\begin{equation} \label{e:GPq} \hat q(x) = \VV(f(x)|Z',X')+\sigma^2 \end{equation}
where the posterior variance is calculated in equation \eqref{e:GPv}. If $(Z,X)$ is a new observation from the model, then the posterior mean and variance of $Z$ given $Z',X',X$ are $\hat f(X)$ and $\hat q(X)$, respectively.
For $d > 1$, we repeated this process independently for each dimension to separately learn the coordinates of $f$.

We learn $Q$ in two different ways. The first method, DKP-GP, estimates $Q(x)$ as a diagonal matrix. The $i$th entry of the diagonal is the corresponding $\hat q(x)$ from \eqref{e:GPq}. The second method, DKP-GP-freq, fixes $\hat f$, and then estimates $Q$ using the covariance of the residuals on a held-out portion (20\%) of the training data. In this case $Q(\cdot)$ is assumed to be constant in $x$ (but not necessarily diagonal).

\subsection{Neural network regression} \label{s:nn}

Artificial neural networks (NN) are another popular method for nonlinear regression because they are universal function approximators. We can learn $f:\RR^m\rightarrow \RR^d$ as a NN by partitioning the data into training, validation, and NN-testing sets, and then using mean squared error (MSE) as an objective function. Thus, our mean estimate $f(\cdot)$ is this neural network, and we estimate $Q(\cdot)=\hat Q$ as the sample covariance of the error. We also used NN to learn $h(\cdot)$ in the EKF and UKF. 

Our implementation consisted of a single hidden layer of 20 tansig neurons trained via Levenberg-Marquardt optimization with Bayesian regularization. 

\section{Model Validation}

\subsection{Synthetic Dataset 1}

To illustrate a simple non-linear, non-Gaussian filtering problem, we specify the following model with $d=1$: 
\begin{align}
\label{e:syn1} Z_t &=  0.9Z_{t-1} + \gamma_t \\
\label{e:syn2} X_{tk} &=  \arctan(Z_t/k) + (\pi\zeta_{tk} + 0.2\theta_{tk}) \quad \quad (k=1,\dotsc,m)
\end{align}
where $\gamma_t,\theta_{tk}\sim^\text{i.i.d.}\N(0,1)$ and $\zeta_{tk}\sim^\text{i.i.d.}\operatorname{Unif}\{-1,0,1\}$.
This model has linear Gaussian state dynamics, but nonlinear and non-Gaussian observations. In fact, the observation noise $(\pi\zeta_{tk} + 0.2\theta_{tk})$ is strongly multimodal, being the sum of normally distributed noise and discrete uniform noise. Because of the multimodal noise, the value of $X_{tk}$ jumps between one of three different arctan clusters. For both of these reasons, the Kalman filter would be predicted to be a suboptimal filter choice. Moreover, the EKF and UKF would be expected to perform suboptimally because the noise in \eqref{e:syn2} is not normal. In contrast, the DKF would be learning $p(z_t|x_t)$, which we would expect to be better approximated by a Gaussian distribution, particularly for large $m$ because of the Bernstein--von Mises phenomenon.

We generated sequences 10,000 points and performed 5 trials in $m=5$ dimensions, splitting the sequences evenly into contiguous training and testing segments. 
\begin{table}[h]
  \caption{Synthetic dataset 1 (normalized MSE)}
  \label{tab:syn1}
  \centering
  \begin{tabular}{l||c|c|c|c|c||c}
  	 filter & trial\#1 & trial\#2 & trial\#3  & trial\#4  & trial\#5  & avg.\\ \hline\hline
     Kalman & 0.565   & 0.561 &   0.546 &    0.561 &    0.510  & \textbf{0.549}\\ \hline
     EKF & 0.643   & 0.636  &  0.624 &   0.698   & 0.597 & \textbf{0.640}\\ \hline
     UKF & 0.595   & 0.578  &  0.576   & 0.594  &  0.521 & \textbf{0.573}\\ \hline \hline
     DKF-GP & 0.062  &  0.066   & 0.071  &  0.075  &  0.087 & \textbf{0.069}\\ \hline
     DFK-GP-freq & 0.066 & 0.072 & 0.075  & 0.084  & 0.075 & \textbf{0.075} \\ \hline
     DKF-NN & 0.074  &  0.103   & 0.090  &  0.117  &  0.087 & \textbf{0.094}\\ \hline
  \end{tabular}
\end{table}
As predicted, there is a difference in normalized MSE values between the generative and discriminative approaches. 



\subsection{Synthetic Dataset 2}

For our second example, we specify the following model, again with $d=1$: 
\begin{align}
\label{e:2syn1} Z_t &=  0.9Z_{t-1} + \gamma_t \\
\label{e:2syn2} X_{t} &=  \big( |Z_t|+ 0.1\theta_{t1}, \ \operatorname{sign}(Z_t) + 0.1\theta_{t2} \big)^\tr
\end{align}
where $\gamma_t,\theta_{tk}\sim^\text{i.i.d.}\N(0,1)$, $k=1,2$.  Note that the relationship between states and observations in this model has the form specified in \eqref{e:EKF-X}.  However, the function $h: \RR\rightarrow \RR^2$ taking $x$ to absolute value and sign is discontinuous and highly nonlinear.  Even if EKF knew the true form of $h$, linearizing about the mean of this model would result in an extremely poor approximation to this model.  Similarly, given the true form of $h$, the discontinuity and nonlinearity of $h$ would disrupt the way the UKF employs the unscented transform to map reference points and estimate mean and variance.  On the other hand, $X_t$ does provide strong information about $Z_t$, so we would expect $p(z_t|x_t)$ to have a dominant mode and to be easily learned from data and incorporated into the DKF.

We generated sequences 2,000 points and performed 5 trials in this $m=2$ dimensional model, splitting the sequences evenly into contiguous training and testing segments. 
\begin{table}[h]
  \caption{Synthetic dataset 2 (normalized MSE)}
  \label{tab:syn2}
  \centering
  \begin{tabular}{l||c|c|c|c|c||c}
  	filter & trial\#1 & trial\#2 & trial\#3  & trial\#4  & trial\#5  & avg.\\ \hline\hline
     Kalman & 0.335  &  0.380  &  0.428  &  0.344  &  0.309  & \textbf{0.359}\\ \hline
     EKF & 1.893  & 11.154  & 8.580  &  2.019  &  1.815 & \textbf{5.092}\\ \hline
     UKF & 1.251  &  2.232  &  2.123   & 3.268 &  13.613 & \textbf{4.498}\\ \hline \hline
     DKF-GP &  0.280  &  0.009  &  0.006   & 0.003  &  0.003 & \textbf{0.060}\\ \hline
     DFK-GP-freq & 0.054  &  0.033   & 0.033   & 0.006  &  0.005 & \textbf{0.026} \\ \hline
     DKF-NN & 0.002   & 0.003  &  0.002  &  0.003  & 0.002 & \textbf{0.002}\\ \hline
  \end{tabular}
\end{table}
We again note the difference in normalized MSE values between the generative and discriminative approaches. 

\subsection{Rhesus Motor Cortex Dataset} \label{s:monkey}

A standard filtering problem in neuroscience consists of decoding continuous movements from neural data in animal models. Data was collected from a microelectrode array chronically implanted into the primary motor cortex of a Rhesus macaque ({\em Macaca multatta}) performing a center-out task \cite{Vargas-Irwin:2015fg} \cite{VargasIrwin:2007ixa}. The monkey moved a planar manipulandum to one of eight different targets that were equally radially distributed: 0, 45, \dots , 315 for a total of 8 targets. The position of the manipulandum was sampled at 50Hz.

A total of 103 spike-sorted neurons were recorded from the NHP that performed 114 different outward movements. Spikes were binned in 100 millisecond intervals, with a lag of 100ms between latent and kinematic variables, simulating neurophysiological parameters. The local Institutional Animal Care and Use Committee approved all experimental procedures. 

We divided the data into 5 non-overlapping sets of 5000 contiguous training points, and then tested on the next 5000 points.  We report normalized MSE (MSE divided by sum of variance in each dimensions of the testpoints).  Table~\ref{tab:monkey_results} summarizes the results.  We find that DKF-GP reduces MSE over the Kalman filter by an average of 27\%; a 2-sided t-test shows the results are highly significant ($p<0.001$).  The DKF-NN reduces MSE over the Kalman by an average of 20\%; similarly, the improvement is highly significant ($p=0.004$).  On the other hand, the EKF and UKF yield results that are indistinguishable from simply predicting the mean.  This might seem surprising, except that implementing them required learning a neural network $h:\RR^2\rightarrow\RR^{103}$.   We see how learning $p(z_t|x_t)$ proved much more tractable than learning $p(x_t|z_t)$ for this problem.

\begin{table}[h]
  \caption{Monkey Neural Filtering Results (normalized MSE)}
  \label{tab:monkey_results}
  \centering
  \begin{tabular}{l||c|c|c|c|c||c}
  	 filter & trial\#1 & trial\#2 & trial\#3  & trial\#4  & trial\#5  & avg.\\ \hline\hline
     Kalman & 0.589   & 0.518 &   0.564 &    0.696 &    0.590  & \textbf{0.591}\\ \hline
     EKF & 0.999   & 1.071  &  0.996 &   1.001   & 1.001 & \textbf{1.013}\\ \hline
     UKF & 3.114   & 1.994  &  2.095   & 1.002  &  0.981 & \textbf{1.837}\\ \hline \hline
     DKF-GP & 0.436  &  0.414   & 0.410  &  0.464  &  0.427 & \textbf{0.430}\\ \hline
     DFK-GP-freq & 0.426 & 0.404 & 0.403  & 0.460  & 0.423 & \textbf{0.423} \\ \hline
     DKF-NN & 0.446  &  0.430   & 0.445  &  0.488  &  0.505 & \textbf{0.463}\\ \hline
  \end{tabular}
\end{table}

\begin{figure}[h]
  \centering
	\fbox{\includegraphics[width=0.6\textwidth]{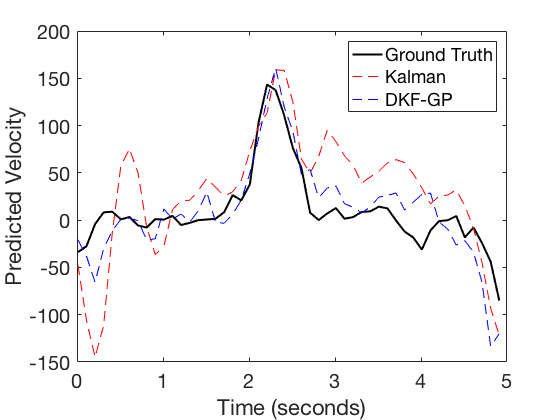}}
  	\caption{Filtering performance comparing the Kalman filter with the DKF-GP with the actual velocity of the manipulandum. In this example, the manipulandum is held stationary over a target for the first two seconds, before the NHP moves to a target and then holds there for another several seconds, before reversing its movement direction}
	\label{fig:p_v_d}
\end{figure}
\section{Discussion}

In this paper, we introduced a discriminative version of the Kalman filter. The DKF provides a fast, analytic filtering solution for models with linear, Gaussian dynamics, but nonlinear, non-Gaussian observations. It requires the models to be specified in an unusual way, but this is not a limitation if the models must be learned from training data. In fact, in situations with higher observation-dimensionality than state-dimensionality, it can be much easier to learn the components needed for the DKF than to learn the ones needed for a more classical filtering approach. The DKF performs well in simulations, and gives state-of-the-art neural decoding results.

The DKF is a special case of a more general strategy of using alternatively-specified models that are both more convenient to learn from data and more convenient to use in filtering applications. We suspect that this strategy could be applied more broadly. Equation \eqref{e:BF2} is one example of an alternative specification, but there are others, such as,
\begin{equation} \label{e:BF3} p(z_t|x_{1:t}) \propto \int \frac{p(z_{t-1}|x_t)}{p(z_{t-1})}p(z_t|z_{t-1},x_t)p(z_{t-1}|x_{1:t-1})\; dz_{t-1} , \end{equation}
which is particularly convenient for particle filtering.
To use equation \eqref{e:BF3} for filtering, one would need to know or learn $p(z_t|z_{t-1},x_t)$ and either $p(z_{t-1}|x_t)$ and $p(z_{t-1})$ or the density ratio $p(z_{t-1}|x_t)/p(z_{t-1})$. 
Monte Carlo methods, like particle filtering, can be used to approximate any of these filtering equations regardless of the model, at least in principle. The DKF assumptions are needed only to ensure an analytic solution. Besides particle filtering, we suspect that many of the methods that have been proposed to generalize the Kalman filter could also be adapted to generalize the DKF.  For instance, if the state dynamics are also nonlinear, then the same ideas underlying the EKF or the UKF should combine with the exact DKF solution to give fast approximations to equation \eqref{e:BF2} for systems with both nonlinear observations and nonlinear dynamics.

\subsection{Comparison to other methods}

Other filtering approaches, incorporating non-parametric regression techniques, have been described in the literature. For instance,  the Gaussian process factor analysis (GPFA) method models $p(x_t|z_t)$ as linear plus Gaussian noise, while $p(z_t)$ is modeled by a 0-mean Gaussian process \cite{Yu:2009ex}. Since the latent variables are not observed, both the filter parameters and latent variables are learned by expectation maximization. This method introduces the machinery of Gaussian processes, but maintains a linear relationship between observations and latent variables.

The GP-BayesFilters uses HMM with the Gaussian process framework~\cite{Ko:2009kg}. Both $p(x_t|z_t)$ and $p(z_t|z_{t-1})$ are modeled by Gaussian processes so that a particle filtering or UKF approximation is required.  The approach is presented for the case that the observations are scalars ($m=1$), and applying the filter to our Rhesus Motor Cortex Dataset for example would require learning 103 different Gaussian processes, one for each observed dimension.  

\subsection{Limitations}

There are a few limitations to the DKF method worth enumerating. First, the DKF assumes a unimodal posterior and is unlikely to work well in problems where it is important to maintain the full shape of a multimodal posterior. This is a limitation shared by the Kalman filter, EKF, and UKF; however, it is not a limitation of the more general strategy of using alternatively-specified filters, since particle filtering (or other methods) can be used to approximate multimodal posteriors.  Second, the DKF is specifically designed to model situations where the dimensionality of the observation space is much greater than that of the latent space. In cases where the converse is true, generative approaches may be better suited for the application. Third, because the DKF has moved away from a generative model, it loses the ability to generate samples from the joint distribution. Finally, it is worth noting that the simple learning methods that we suggest in this paper completely ignore the temporal structure of the training data and could be significantly improved, although likely at the expense of creating custom training software. 
 
 Implementing the DKF requires learning the model for $p(z_t|x_t)$. While our implementation of the DKF-GP and DKF-NN performed very well (despite the 103 input dimensions of neural data!), one could expect that using non-parametric regression techniques would encounter difficulties when trained with high-dimensional data. 
 
 \subsection{Future work}
 
In situations where the DKF is applicable, the combination of speed, flexibility, and easy of use set it apart. Because of this, in future work we intend to explore the use of the DKF as an approximation algorithm in situations where the generative state-space model is known, but the classical filtering recursion in equation \eqref{e:BF} is too time consuming to approximate with existing techniques.
 
 \small
\bibliographystyle{plain}
\bibliography{nips_2016}

\begin{thebibliography}{10}

\bibitem{Abb:2005}
P.~Abbeel, A.~Coates, M.~Montemerlo, A.~Y. Ng, and S.~Thrun.
\newblock Discriminative training of kalman filters.
\newblock In {\em Proc Robotics: Science and Systems}, 2005.

\bibitem{Vargas-Irwin:2015fg}
C.~Vargas-Irwin, D.M.~Brandman, J.B.~Zimmermann, J.P.~Donoghue, M.J.~Black
\newblock Spike Train SIMilarity Space (SSIMS): A Framework for Single Neuron and Ensemble Data Analysis.
\newblock {\em Neural Computation}, 27(1):1--31,2015.

\bibitem{VargasIrwin:2007ixa}
C.~Vargas-Irwin, J.P.~Donoghue
\newblock Automated spike sorting using density grid contour clustering and subtractive waveform decomposition.
\newblock {\em Journal of Neuroscience Methods} 164:1--18,2007.

\bibitem{chen2003bayesian}
Zhe Chen.
\newblock Bayesian filtering: From kalman filters to particle filters, and
  beyond.
\newblock {\em Statistics}, 182(1):1--69, 2003.

\bibitem{Doucet:2000}
A.~Doucet, S.~Godsill, and C.~Andrieu.
\newblock On sequential {M}onte {C}arlo sampling methods for {B}ayesian
  filtering.
\newblock {\em Statistics and Computing}, 10(3):197--208, 2000.

\bibitem{Gordon:1993}
N.~J. Gordon, D.~J. Salmond, and A.~F.~M. Smith.
\newblock Novel approach to nonlinear/non-{G}aussian {B}ayesian state
  estimation.
\newblock {\em Radar and Signal Processing, {IEE} Proc F}, 140(2):107--113,
  Apr. 1993.

\bibitem{hess2009discriminatively}
Robin Hess and Alan Fern.
\newblock Discriminatively trained particle filters for complex multi-object
  tracking.
\newblock In {\em Computer Vision and Pattern Recognition, 2009. CVPR 2009.
  IEEE Conference on}, pages 240--247. IEEE, 2009.

\bibitem{Julier:1997}
S.~J. Julier and J.~K. Uhlmann.
\newblock New extension of the {K}alman filter to nonlinear systems.
\newblock {\em Proc SPIE}, 3068:182--193, 1997.

\bibitem{Kanazawa:1995}
K.~Kanazawa, D.~Koller, and S.~Russell.
\newblock Stochastic simulation algorithms for dynamic probabilistic networks.
\newblock In {\em Proc {UAI} '95}, pages 346--351, 1995.

\bibitem{Ko:2009kg}
J.~Ko and D.~Fox.
\newblock {GP}-{B}ayes{F}ilters: {B}ayesian filtering using {G}aussian process
  prediction and observation models.
\newblock {\em Autonomous Robots}, 27(1):75--90, 2009.

\bibitem{Ras:2006}
C.~E. Rasmussen and C.~K.~I. Williams.
\newblock {\em Gaussian processes for machine learning}.
\newblock MIT Press, Cambridge, MA, 2006.

\bibitem{Yu:2009ex}
B.~M. Yu, J.~P. Cunningham, G.~Santhanam, S.~I. Ryu, K.~V. Shenoy, and
  M.~Sahani.
\newblock Gaussian-process factor analysis for low-dimensional single-trial
  analysis of neural population activity.
\newblock {\em J Neurophysiol}, 102(1):614--635, Jul. 2009.

\end{thebibliography}

\end{document}